\documentclass{article}

    \usepackage[preprint]{neurips_2020}

\usepackage[utf8]{inputenc} 
\usepackage[T1]{fontenc}    
\usepackage{hyperref}       
\usepackage{url}            
\usepackage{booktabs}       
\usepackage{amsfonts}       
\usepackage{nicefrac}       
\usepackage{microtype}      
\usepackage{graphicx}
\usepackage{natbib}
\usepackage{xcolor}

\usepackage[compact]{titlesec}
\title{Deep Learning on Real Geophysical Data:
A Case Study for Distributed Acoustic Sensing Research}

\author{
  Vincent Dumont, Ver\'{o}nica Rodr\'{i}guez Tribaldos, Jonathan Ajo-Franklin,
  and
  Kesheng Wu\\
  Lawrence Berkeley National Laboratory\\
  \texttt{\{vdumont$\mid$vrodrigueztribaldos$\mid$JBAjo-Franklin$\mid$kwu\}@lbl.gov} 
}

\begin{document}

\maketitle

\begin{abstract}
Deep Learning approaches for real, large, and complex scientific data sets can be very challenging to design. In this work, we present a complete search for a finely-tuned and efficiently scaled deep learning classifier to identify usable energy from seismic data acquired using Distributed Acoustic Sensing (DAS). While using only a subset of labeled images during training, we were able to identify suitable models that can be accurately generalized to unknown signal patterns. We show that by using 16 times more GPUs, we can increase the training speed by more than two orders of magnitude on a 50,000-image data set. 
\end{abstract}

\section{Introduction}
The increasing occurrence of large and complex data sets has forced the scientific community to explore innovative approaches to enable efficient data analysis. In this context, the combination of High-Performance Computing (HPC) systems and Machine Learning (ML) algorithms represents an attractive approach that promises to leverage this new challenge. In this work, we use data from a novel sensing technology called Distributed Acoustic Sensing (DAS), which utilizes optical fibers to record seismic activity at spatial density of a few meters for distances of 10's of km \citep{Ajo-Franklin2019}. Such high-density array measurements can amount to several Terabytes of data within a week, making them challenging to comprehensively analyze using standard geophysical data analysis tools. However, the high density of spatial sensors and high temporal sampling made available through the DAS technology allows geophysicists to visualize seismic events in the form of 2-dimensional spatiotemporal images, thereby making the data suitable for exploring image classification techniques using machine learning.

The ultimate objective of our method is to identify coherent, usable seismic energy within noisy DAS data sets, and assign a probability of occurrence of these signals to each data record. However, the types of signals present in DAS data sets can be varied and complex. That's specially true for DAS recordings of so-called ambient seismic noise, where vibrations from natural (e.g. wind, the ocean) and anthropogenic (e.g. vehicles) sources are acquired to be analyzed for subsurface exploration. In most of these data sets, coherent seismic signals usable for geophysical analysis are embedded in background and instrumental noise, and seismic signals from different sources can interfere, generating complicated signal patterns when visualized as 2-dimensional images. Thus, training a classifier that can efficiently assign the true probability of a particular signal type within a region that is contaminated by other unknown signals can be very challenging. In this work, we introduce a technique to identify suitable models that can be generalized on large volume data sets containing highly complex, non-trained signal patterns using a deep learning approach implemented on a HPC system. The key contributions of this work are: (1) a hyperparameter exploration methodology to test all possible training approaches, (2) a training throughput mapping assessment to determine the right GPU scaling to use at large data set sizes, and (3) a model interpretability approach to identify trustworthy models during inference.
 
\section{Tuning and Hyperparameter Exploration}
\begin{figure}
\includegraphics[width=\columnwidth]{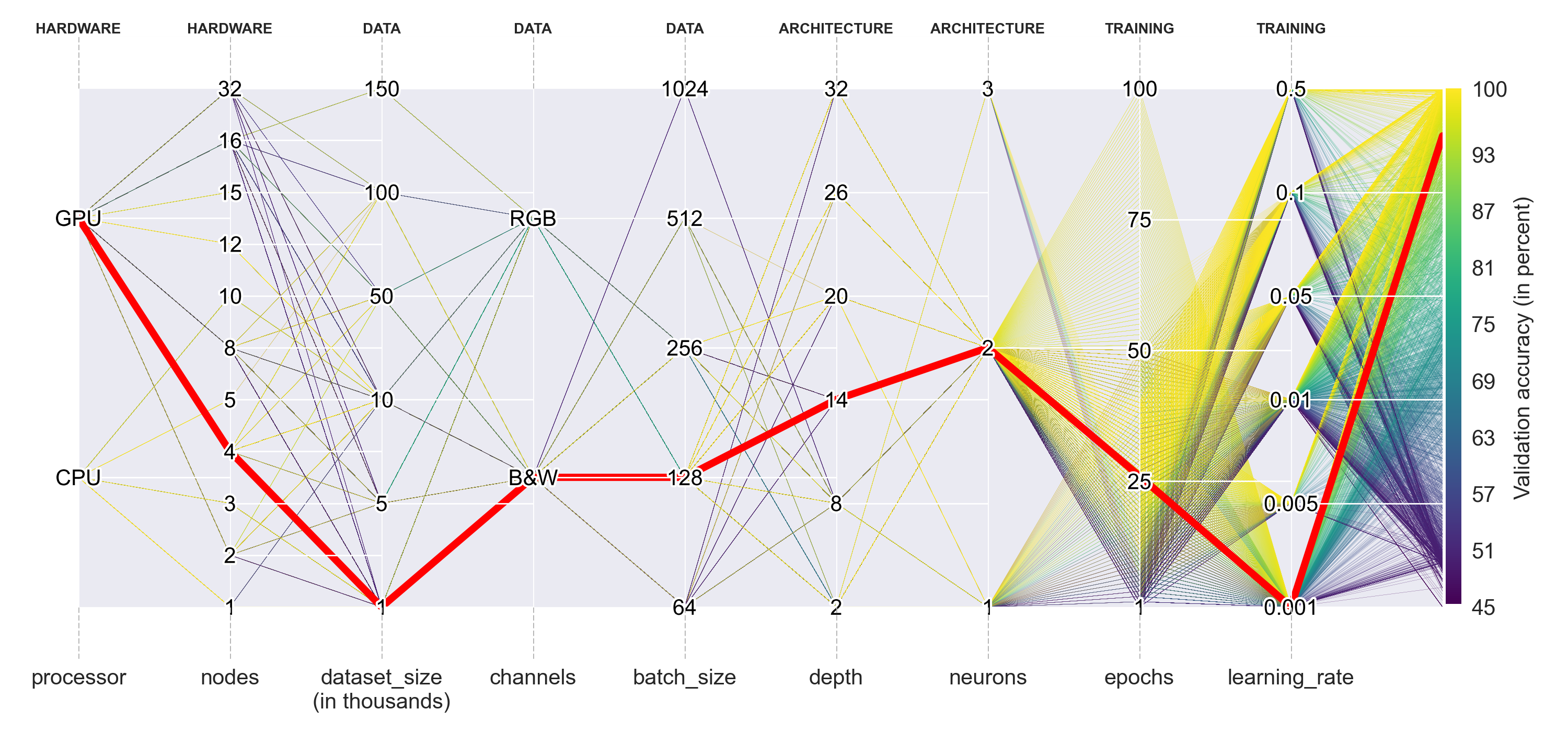}
\caption{Parallel coordinates plot showing all the combinations of parameters explored in this work. The red path highlights the best trained model with results closest to the ground truth.}
\label{parallel_coordinates}
\end{figure}

The tuning was performed using the Cori supercomputer managed by the National Energy Research Scientific Computing Center (NERSC) at the Lawrence Berkeley National Laboratory. A 9-dimensional parameter space was explored where each parameter evaluates different aspects of the training, including HPC hardware, data characteristics and neural network architecture. In Figure \ref{parallel_coordinates}, we summarize all the combinations of parameters that we have explored in this work in the form of a coordinates plot \citep{1997_inselberg}. A total of 6,460 different trained models are presented.

\begin{enumerate}
\item {\bf Hardware:} While the hardware does not ultimately affect the actual outcome of the training, it does impact its speed. It is therefore critical to assess which setup is most efficient when dealing with Big Data such as that explored here. Both Central Processing Units (CPUs) and Graphical Processing Units (GPUs) were independently tested. Up to 10 Haswell processor nodes were used for CPU parallelization for a total of 320 individual cores. While over 2,000 Haswell nodes are available in the Cori cluster, we privileged the use of GPUs for larger distributed setups. GPUs have shown to outperform purely CPU-based parallelization setups in solving deep learning problems \citep{gpu}. In this work, up to 32 individual NVIDIA V100 Volta GPUs were used across 4 GPU nodes.
\item {\bf Data:} While the original raw data can intuitively be thought and treated as black-and-white (B\&W) images, convolutional neural networks often interpret and distinguish features within the input images not only based on the changes in amplitudes but also in colors. Given the complex nature of the DAS data signals, we have explored both colored (RGB) and grayscale (B\&W) images. For colored images, the data were converted into 3-channel tensors, while grayscale image data only have one channel. Apart from the number of channels, different batch sizes are explored, from 64 to up to 512 data records per model training iteration.
\item {\bf Architecture:} In this work, we use a Residual neural Network architecture (ResNet, \citet{resnet}) to build the classifier. While such architecture has proven to be extremely efficient for image classification problems, its depth (i.e. number of hidden layers) and size of bottleneck layer needs to be fine-tuned in order to obtain good classification models. Deeper networks have tendencies to extract finer features of the input image while shallower networks tend to identify larger patterns. Different number of neurons in the final layer where also used to explore different classification approaches, from unary to multiclass approaches.
\item {\bf Training:} The learning rate and number of epochs are two important parameters that will lead the models towards different outcomes. Both parameters work together, where more epochs are usually needed when using smaller learning rates in order to get a good model. However, while optimized models might be achieved faster using large learning rates, such models may sometimes overestimate the probability of some classes.
\end{enumerate}

\section{Scalability and training throughput efficiency}

Given the high density of hyperparameter combinations tested during the tuning, it is critical to build highly scalable algorithms. Three parameters from which the training throughput (that is, the speed at which the training is completed) is highly dependent on can be identified, namely the depth of the neural network, the batch size used for each training iteration, and the size of the input data set used. The impact of using different neural network depths is straightforward to assess, where a combination of large node usage with shallower neural networks will consistently provide the highest training throughput. However, this outcome does not hold true for neither data set nor batch size parameters.

\begin{figure}
\includegraphics[width=\columnwidth]{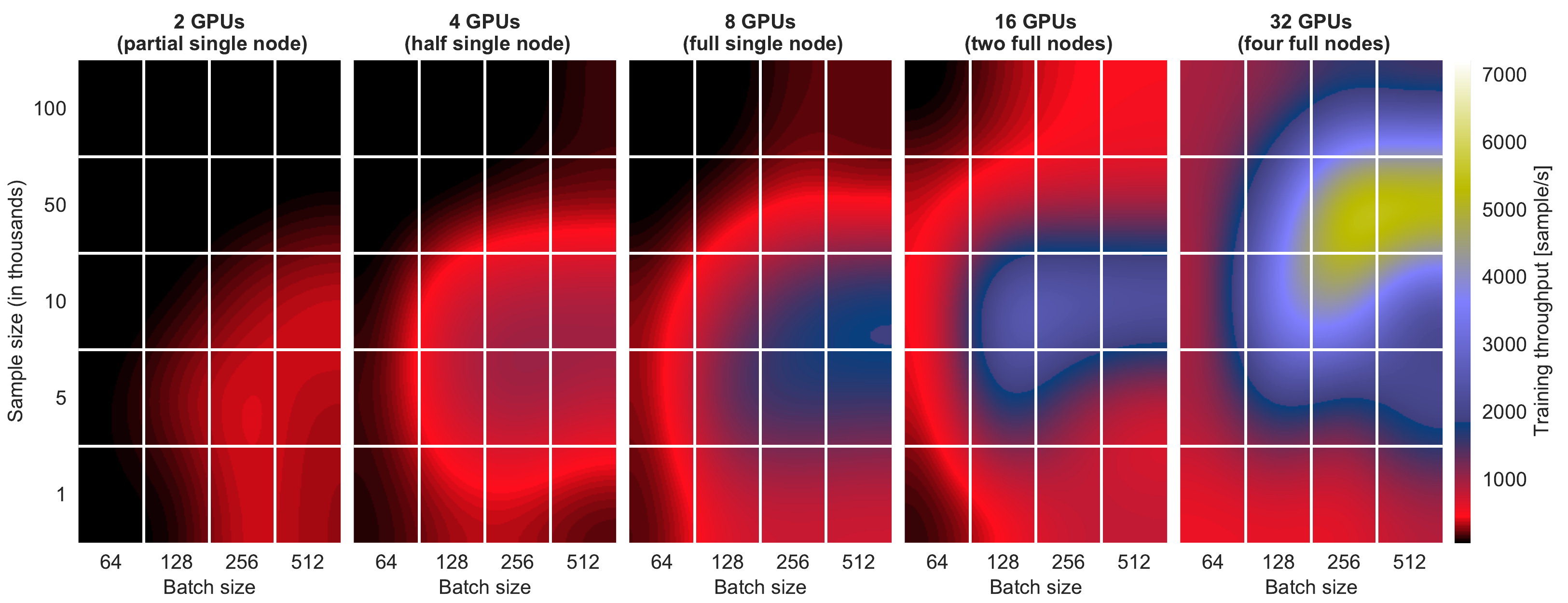}
\caption{Evolution of training throughput efficiency on a single epoch across different GPU scaling. Higher training throughput values correspond to faster training. This figure demonstrates how larger scale distributed training is most efficient on larger data set sizes, whereas smaller distributions are more adequate for smaller data sets. Up to two orders of magnitude improvement in speed were able to be achieved by using the right parameter combination at high GPU scaling.}
\label{gpu_scaling}
\end{figure}

The combination of data set size and batch size providing the fastest training will change with the number of GPUs used, as shown in Figure \ref{gpu_scaling}. For instance, the training speed will be at his highest if using a 1,000-image data set across 2 GPUs, whereas for a 32-GPU setting, the highest throughput will be reached if using a larger, 50,000-image data set. In fact, we were able to improve the throughput by more than two orders of magnitude on a 50,000-image data set trained with a batch size of 512. Indeed, a training speed of 65 sample/second was achieved using 2 GPUs while we managed to reach a throughput of 7,210 sample/second when using a 32 GPUs distributed setup.

\section{Interpretability and Model Validation}
As mentioned above, we aim at building an image classifier that can estimate how much usable energy is present in a given DAS data region. Such energy signals are very common in regions where ambient seismic vibrations, mostly from human origin, are recorded. While signals with usable energy are typically represented as coherent seismic waves, they can, however, take more complex forms. For instance, in regions with high levels of ambient seismic noise, interfering coherent waves will appear as complex patterns that are hard to evaluate in 2-dimensional images. Similarly, regions where energies are so high that saturate the recorder will become unusable, but will be surrounded with coherent energy.

\begin{figure}[ht]
\includegraphics[width=\columnwidth]{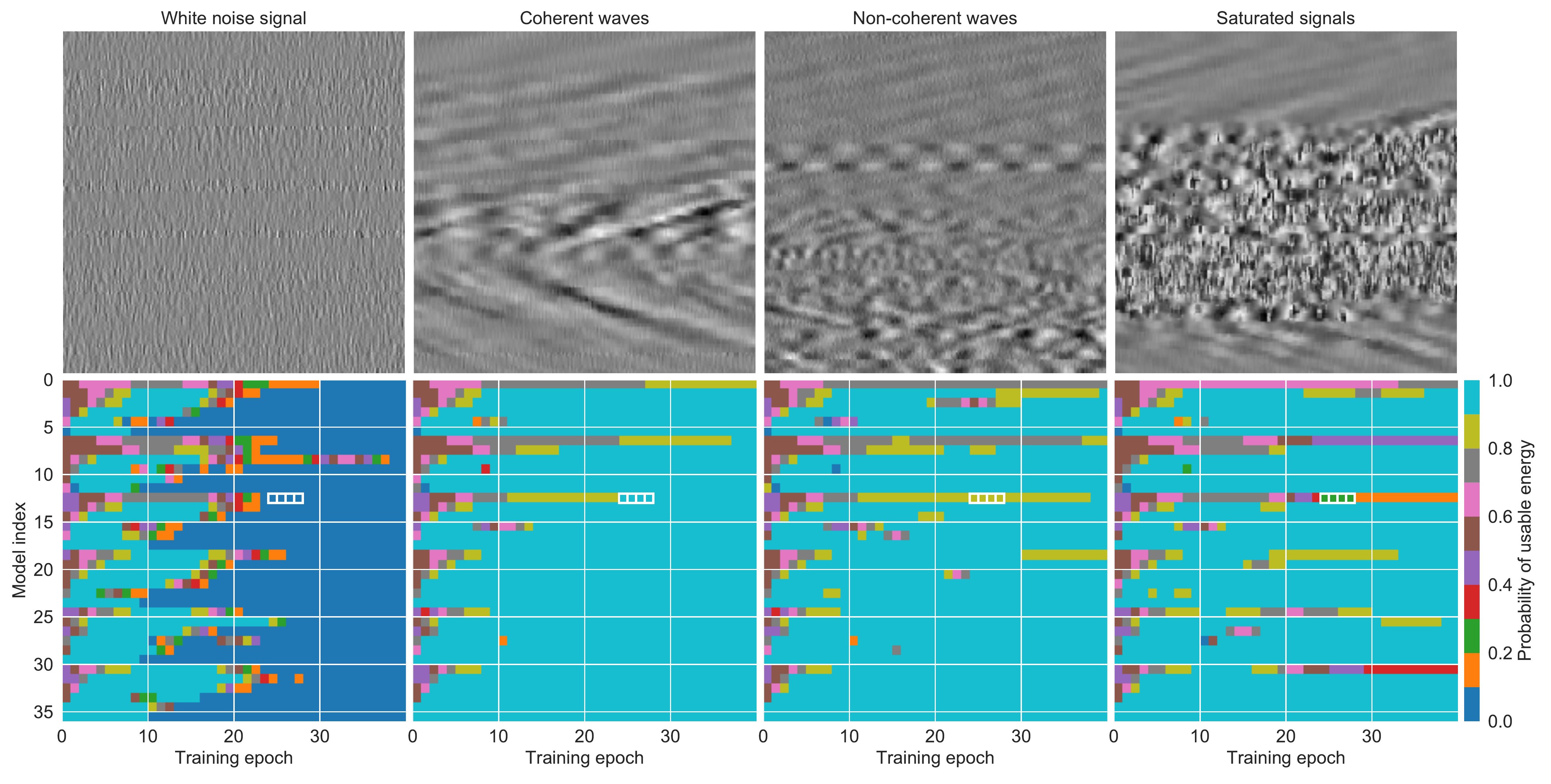}
\caption{Evolution of the probability of usable energy for 4 reference signal types across 36 different models and 40 consecutive epochs. Each model was trained on a 1,000-grayscale-image data set using 4 GPUs, a batch size of 128 and 2 neurons in bottleneck layer. The model indexes correspond, in order, to depths ranging from 2 to 32 hidden layers and learning rates from 0.001 to 0.5 for each explored depth. The white squares indicate the coordinates where probabilities of usable energy are closest to the ground truth.}
\label{prob_assess}
\end{figure}

The top panels from Figure \ref{prob_assess} show the four reference types of signal patterns that can be found in the DAS data set analyzed in this work. Of these four types, only white noise signals and coherent waves can be used to build a training data set. The complexity of non-coherent wave patterns is such that there is no good metric to identify such region in a satisfactory way. As for regions with saturated signals, their low frequency of occurrence make them unsuitable for building large data sets. Our classifier therefore works in binary mode where any given input image will be identified either as coherent surface waves or white noise signals. This, however, presents an issue for non-coherent and saturated signal regions where the expected probability of usable energy might be overestimated.

The bottom panels from Figure \ref{prob_assess} show how the output probability of usable energy changes for 36 different models over 40 consecutive epochs. Identifying which models output a probability that is consistent with what we expect is very challenging and one will still rely on some human expertise to evaluate the quality of each trained model. In Table \ref{best_models}, we show the expected probability for usable surface wave energies for each reference type of signal shown in Figure \ref{prob_assess}. Across all 1,440 trained models, only 4 models were found to match the expected probabilities. These 4 models have shown to provide highly accurate probabilities for usable energy across an entire 10-day period of recorded DAS data.

\renewcommand{\arraystretch}{1.5}
\begin{table}
\centering
\caption{Ground truth and deep learning model output probabilities for the 4 best trained models. The best models correspond to 4 consecutive epochs of a 14-hidden layer deep model trained with a batch size of 128 and learning rate of 0.001 on a 1,000-image data set.}
\label{best_models}
\begin{tabular}{l c c c c c}
 \hline
 {\bf Signal type} & {\bf Expected range} & {\bf Epoch 25} & {\bf Epoch 26} & {\bf Epoch 27} & {\bf Epoch 28} \\
 \hline\hline
 White noise & $p<0.1$ & 0.0657 & 0.0449 & 0.0320 & 0.0237 \\
 \hline
 Coherent waves & $p>0.9$ & 0.9008 & 0.9058 & 0.9111 & 0.9163\\
 \hline
 Non-coherent waves & $0.7<p<0.9$ & 0.8551 & 0.8566 & 0.8592 & 0.8628 \\
 \hline
 Saturated signal & $0.2<p<0.3$ & 0.2975 &  0.2568 & 0.2268 & 0.2048\\
 \hline
\end{tabular}
\end{table}

\section{Conclusion}
We presented a complete hyperparameter exploration to build highly-scalable machine learning classifier models designed to identify usable seismic energy signals in Distributed Acoustic Sensing (DAS) data.  While the models were not trained on contaminated (i.e., non-coherent, saturated) data regions, we are able to select the best classifier among thousands of trained models using the expected probability range on only four reference images. Finally, we developed a GPU scaling assessment method that allowed us to improve by up to two orders of magnitude the training throughput on large training data sets.

\section*{Broader Impact}
Our study provides an approach to evaluate the performance of a deep learning image classifier on HPC.
We applied this classifier on a large, complex Distributed Acoustic Sensing (DAS) dataset.
DAS technology is revolutionizing the acquisition of seismic data, however it is generating too much data for existing analysis techniques.
We are applying Machine Learning approaches to DAS data and studying issues such as scalability, interpretability and other issues important to the domain scientists.
The analysis techniques we developed could help advance research in a variety of topics with great societal impact, such as real-time imaging natural and induced seismic events from oil and gas reservoirs, hydraulic fracturing, geothermal energy systems,
urban traffic, critical infrastructure (such as highways and oil pipelines) among many others.
Our work on scalability and interpretability of ML techniques is also applicable to other scientific applications that extract complex signals from large data collections.  

\begin{ack}
This work is supported by the Laboratory Directed Research and Development (LDRD) Program of Lawrence Berkeley National Laboratory under U.S. Department of Energy Contract No.~DE-AC02-05CH11231. 
It used resources of the National Energy Research Scientific Computing Center and Energy Science network (ESnet), both are funded under the above contract.
\end{ack}

\bibliography{references}

\end{document}